\documentclass{article}

\usepackage[preprint]{neurips_2022}

\usepackage{amsmath,amsfonts,bm}

\def\eqref#1{equation~\ref{#1}}

\def\1{\bm{1}}

\DeclareMathAlphabet{\mathsfit}{\encodingdefault}{\sfdefault}{m}{sl}
\SetMathAlphabet{\mathsfit}{bold}{\encodingdefault}{\sfdefault}{bx}{n}

\usepackage[utf8]{inputenc} 
\usepackage[T1]{fontenc}    
\usepackage{hyperref}       
\usepackage{url}            
\usepackage{booktabs}       
\usepackage{amsfonts}       
\usepackage{nicefrac}       
\usepackage{microtype}      
\usepackage{xcolor}         
\usepackage{subfig}

\usepackage{algpseudocode}
\usepackage{algorithm}
\usepackage{graphicx}

\usepackage{listings}
\usepackage{xcolor}

\DeclareUnicodeCharacter{2217}{*}

\title{From Language Models to Practical Self-Improving Computer Agents}

\author{%
  Alex Sheng \\
  New York University \\
  \texttt{alexsheng4@gmail.com} \\
}

\begin{document}

\maketitle

\begin{abstract}
We develop a simple and straightforward methodology to create AI computer agents that can carry out diverse computer tasks and self-improve by developing tools and augmentations to enable themselves to solve increasingly complex tasks. As large language models (LLMs) have been shown to benefit from non-parametric augmentations, a significant body of recent work has focused on developing software that augments LLMs with various capabilities. Rather than manually developing static software to augment LLMs through human engineering effort, we propose that an LLM agent can systematically generate software to augment itself. We show, through a few case studies, that a minimal querying loop with appropriate prompt engineering allows an LLM to generate and use various augmentations, freely extending its own capabilities to carry out real-world computer tasks. Starting with only terminal access, we prompt an LLM agent to augment itself with retrieval, internet search, web navigation, and text editor capabilities. The agent effectively uses these various tools to solve problems including automated software development and web-based tasks.
\end{abstract}

\section{Introduction}

Large Language Models (LLMs) pretrained on internet-scale data have been shown to develop some degree of emergent reasoning capabilities. This has enabled them to interact with tools and augmentations comprised of external software that interfaces with LLMs through their input and output pipelines. These pieces of external software may implement prompt engineering techniques, may be non-parametric, and have been associated with various different terminologies including "augmentations," "tools," and "plugins." We view these various types of model-augmenting software to be expressions of the same overarching concept of model augmentation, which this paper aims to better understand and systematize. We introduce a practical methodology to automate the development of model-augmenting software by creating agents that can autonomously produce tools and augmentations to suit their own needs. These self-improving agents can then be used to flexibly address diverse computer tasks, generating software to augment themselves and complete complex tasks that they are initially unable to solve.

The following section provides an overview of the current state of research in large language models, language model augmentations, and self-improving AI, highlighting the key contributions and developments in each area.

\subsection{Large Language Models}

\textbf{Foundation Models.} The development of large language models has been a significant area of research, with the Transformer architecture \citep{attention} serving as the foundation for many subsequent models \citep{gpt3,gpt4}. The scaling laws for neural language models \citep{scaling} have been explored to understand how model performance scales with size. Foundation models \citep{foundation_models} have been discussed as a new paradigm in AI, providing a shared basis for various downstream tasks \citep{instruct}.

\textbf{Code Generation.} Code generation has been a notable application of large language models, with Codex \citep{codex} demonstrating the ability to generate code from natural language descriptions. The AlphaCode system \citep{alphacode} further pushed the boundaries by competing in programming competitions. CodeRL \citep{coderl} explored code generation through deep reinforcement learning.

\textbf{Language Model Agents.} Language models have been employed as agents in decision-making tasks \citep{decision_making}, with models like Reflexion \citep{reflexion} using verbal reinforcement learning to improve performance. InterCode \citep{intercode} presents a framework for interactive coding with execution feedback as a reinforcement learning environment, providing a foundation for our environment design.

\subsection{Language Model Augmentations}

\textbf{Retrieval.} Starting with early works on parametric memory \citep{ntm,mann}, memory retrieval has proven to be a flexible and effective way to augment deep learning models to address knowledge-intensive tasks. Retrieval-augmented generation models like RAG \citep{rag} have combined language models with external knowledge retrieval to enhance their capabilities in knowledge-intensive tasks. REALM \citep{realm} and kNN-LM \citep{knnlm} are other notable examples in this area.

\textbf{Reasoning.} Approaches such as Chain-of-Thought prompting \citep{cot} have been explored to elicit reasoning in large language models. Rethinking with Retrieval \citep{retrieval_cot} and Interleaving Retrieval with Chain-of-Thought Reasoning \citep{ircot} have been investigated for multi-step questions.

\textbf{Internet.} Internet-augmented language models \citep{internet_augmented} and systems like WebGPT \citep{webgpt} and WebShop \citep{webshop} have been developed to enable models to interact with the web for open-domain question answering.

\textbf{Program Evaluation.} The evaluation of code generated by language models has been a focus area, with approaches like PAL (Program-aided Language Models) \citep{pal} being explored to enhance language models with programming capabilities. Program of Thoughts Prompting \citep{program_of_thoughts} has been investigated as a method to disentangle computation from reasoning in numerical reasoning tasks, demonstrating the potential for language models to perform more sophisticated code evaluations.

\textbf{Augmented Language Model Agents.} Augmented language models have been used to create more capable agents, as seen in works like Seeker \citep{seeker}, ReAct \citep{react}, and HuggingGPT \citep{hugginggpt}, which synergize reasoning and acting. Toolformer \citep{toolformer} demonstrated language models teaching themselves to use tools. The LaMDA model \citep{lamda} and BlenderBot 3 \citep{blenderbot} have been developed for dialog applications with continual learning capabilities.

\subsection{Self-Improving AI}

Self-improving AI systems have been a topic of long-standing interest, with the Godel Machine \citep{godel} being a prominent theoretical framework for a self-referential universal problem solver. Self-Referential Learning \citep{self_referential} and the Self-Taught Optimizer (STOP) \citep{stop} have further explored recursive self-improvement.

\section{Methods}

\begin{algorithm}[hbt!]
\caption{ Algorithm}\label{alg:wordy}
\begin{algorithmic}
\Require{Language model query endpoint}
\Ensure{Implement a helpful computer agent}
    \State $snippets$ is an accessible location within the environment that stores code
    \State $query() \gets$ language model query endpoint
    \State $prompt \gets$ environment instructions
    \While{agent is active}
        \State $task \gets$ wait for human input to define a task
        \State $prompt \gets prompt + task$
        \While{\textbf{not} (task completion condition \textbf{or} max steps reached)}
            \State $response \gets query(prompt)$
            \State $prompt \gets$ empty prompt template
            \If{$response$ contains code blocks}
                \State $snippets \gets$ code blocks parsed from $response$
            \EndIf
            \If{$response$ contains terminal commands}
                \State $commands \gets$ terminal commands parsed from $response$
                \State $output \gets$ empty output template
                \For{$command$ in $commands$}
                    \State $output \gets output + execute(commands)$)
                \EndFor
                \State $prompt \gets output$
            \EndIf
            \If{$response$ requests human input}
                \State \textbf{break}
            \EndIf
        \EndWhile
    \EndWhile
    \label{algorithm:deepphoenix_algorithm}
\end{algorithmic}
\end{algorithm}

We present a methodology for building and operating LLM-powered general computer agents. These agents have the ability to self-improve by generating tools to augment themselves.

\subsection{System Architecture}

The system consists of an instruction-tuned LLM agent capable of generating code and executing terminal commands within a computer environment. This LLM agent can either be a remote model queried through an API or a locally hosted model; the difference is not important, as only the ability to programmatically query this model is necessary. In our experiments, we use a GPT model through the OpenAI API.

\subsection{Prompt Loop}

To operate this system, we implement an algorithmic loop that continuously queries the model. An initial set of instructions is provided with the first prompt, describing the system (including environment information and agent controls) to the model. The model receives human input defining a task which it will generate code and actions (in the form of terminal commands) to complete. Once a task is completed or some kind of loop-limiting criterion is triggered, the system will await additional human input to define another task or otherwise provide information or instructions for the model.

\subsection{Output Parsing and Execution}

Each time the model responds to a query, its output is parsed systematically. The system identifies generated "code blocks" that are separated from other output by delimiters, such as triple backquotes in the case of the GPT API. Generated code blocks can either be actual code or terminal commands, the two of which are treated differently:

\textbf{Generated Code:}
For most programming languages of interest, these parsed sequences of generated code are automatically stored within the computer environment in an accessible place such as a file with a predetermined file path.

\textbf{Terminal Commands:}
For code blocks labeled as terminal commands (for example "bash," "cmd," and "powershell" for the GPT API), these commands are parsed and directly executed within the computer environment. If both code and terminal commands are generated in the same output, then the code is placed into the environment first before terminal commands are executed. This way, executed terminal commands will be able to act on code generated in the same output.

Terminal command execution allows the agent to freely interact with the computer environment. The parsing of generated code to store within the environment is also an important abstraction to help the agent develop software within the environment. For example, the agent can generate code for Python programs, then use the terminal to copy this code into an appropriate program file and execute it.

\textbf{The key insight of this paper} is that an LLM agent within a computer environment can use its code generation abilities to develop software to augment its own capabilities. By using these self-developed tools, the agent can extend its own capabilities to solve more complex tasks within its environment, including potentially creating more sophisticated tools to further extend its capabilities. The result is that we can employ a recursively self-improving LLM agent that can naturally interact with a real-world computer environment and carry out diverse computer tasks. To achieve this, we combine mindfully designed algorithmic and environment components with innovative prompt engineering.

In our experiments, we explore case studies showing how a self-improving agent develops tools to augment itself with a human instructor in the loop. The agent can first generate basic tools for navigating, viewing, and editing software within its environment. With these basic tools for assisting in software development, it can build more complex tools including retrieval augmentation or internet search programs. As more sophisticated tools further increase the agent's ability to access, process, and apply knowledge, it is able to solve increasingly complex tasks.

\section{Experiments}

We carry out three case studies to demonstrate the flexible functionality of our self-improving AI agent. These cases provide, as a starting point, a few examples of what can be achieved with our methodology. These cases represent a very small subset of short, easily demonstrable use cases, and are not at all representative of the full range of possibilities of what can be achieved with self-improving AI agents. To this end, we enthusiastically invite future work to explore new applications. The cases are as follows:

\begin{enumerate}
  \item In the first case, our agent is tasked with creating simple file view and file edit tools which can later be used for software development tasks.
  \item The second case addresses retrieval augmentation methods, showing how our agent is able to autonomously develop rudimentary retrieval augmentation.
  \item The third case demonstrates how our agent can collaborate with a human user and carry out multi-step plans, creating internet search and retrieval tools augmenting itself with internet access to solve web-based tasks.
\end{enumerate}

In our experiments, we use the \textbf{gpt-4-1106-preview} model through the OpenAI API. We focus on evaluating software development tasks in Python in a Windows 11 environment and thus only implement code output parsing for Python code blocks and CMD code blocks. Our agent is given tasks via human user input, and is repeatedly queried without step count restrictions until it generates an output without any code or terminal commands, at which point the system waits for another human user input.

These experiments are provided primarily in the style of tutorials to demonstrate the flexibility and ease of use of self-improving agent workflows, and may represent a "cherry-picked" subset of relatively error-free and easily comprehensible agent outputs. This paper was written to showcase our progress and spark interest among the research community in an interesting new direction with no foreseeable plans to submit for conference publication. Thus, these case studies are designed for ease of comprehension as a first priority, and are not experimentally rigorous to the same standard as similar research with more comprehensive experimental methodology. We have edited certain content to omit sensitive information, significantly shorten message content, and enhance readability.

Listing 1 shows a "system prompt" containing the core instructions that we use to guide self-improving agent behavior in our experiments.

\begin{lstlisting}[label={lst:instructions},caption={System Prompt}]
You are an LLM agent interacting with a computer environment that executes code and terminal commands. The operating system is Windows 11.

Assume that Python is already installed and the "python" terminal command is available. However, you can only see and generate text/language so you can't interact with a text editor interface like a human can.

Suppose your environment provides helper functionality so that you can output code blocks in the format that is natural for you as GPT:

```python
print("Hello, World!")
print("I am alive!")
```

```cmd
echo hello
echo "hello again"
```

The environment will parse Python code blocks you output into the file python_1.txt so that you can more easily write multi-line Python code and then access it from the terminal or within scripts. You can write Python programs that interact with your own generated Python code by simply loading the contents of the python_1.txt file. The name of this file will not change, so you don't have to pass this file name as an argument when executing such programs.

The environment will parse terminal command blocks and execute them as terminal commands. 

You will continuously be re-prompted. If you previously ran a command or executed a script, then the output of your previous command or executed program will be appended to the input, so you can see the results from your previous actions! You can use these outputs to make sure your intended actions were carried out without issues, or to look at output for debugging purposes.

You are interacting with an automatic prompting program. Your responses are parsed programmatically, so please adhere strictly to the following format instructions:
1. Each time you generate an output, the first Python code block in your output will be parsed and stored in python_1.txt.
2. Each time you generate an output, any terminal command blocks in your output will be parsed and executed.
3. If you generate both a Python code block and a terminal command block, the Python code block will first be stored and then the terminal command block will be executed, so that any commands or executed programs will be able to access the Python code through python_1.txt.
4. If you did not generate a Python code block, then the python_1.txt file will remain unchanged, and it will retain the same code content from the last time you did generate a Python code block.
5. You may generate multiple lines of terminal commands within one terminal command block in your output. Terminal command blocks will be executed line by line. Do NOT use operators like "&&" because version support is inconsistent.

You are free to generate textual responses as you normally would, apart from the code blocks which will be parsed systematically. It might be helpful to talk through things step-by-step or plan things out. Your responses may or may not be visible to a human in the loop, but you should still operate independently with the assumption that you won't receive any outside help.

Carefully follow these additional rules:
- Do NOT try to manually echo Python code line by line through the command line. You should use commands like "copy" with python_1.txt to efficiently use your Python code as input for terminal actions. You can also create your own tools for creating and editing scripts.
- Do NOT use operators like "&&" which are version-dependent.
- Do NOT generate "examples" of terminal commands. Remember that your outputs are parsed systematically and automatically executed as is without human intervention, so do not ever generate terminal commands that aren't meant to be directly executed.
- Do NOT generate "lazy" code with comments like "# rest of the code goes here." Your code will be systematically parsed without human intervention, so if you do not generate complete, syntactically correct code, it will not work.
- No output from a command does NOT necessarily mean it did not run properly. It means that there was no output.

Now is when the really cool part starts. Knowing these things, you can try to make tools for yourself to use within this environment, in the form of reusable programs! Take advantage of all the features of your environment, which I will review for you below:
- You can freely run terminal commands. You can generate multiple terminal commands per response. I recommend doing this to minimize repeated queries.
- You can generate natural GPT-style Python code blocks which will automatically be stored into python_1.txt to be accessible within the environment through terminal commands or within scripts.
- You will continuously be re-prompted with the output of your previous action(s)

Just to point you in the right direction, you can output a Python block and then copy from python_1.txt into a .py file. You can create reusable programs that can be called from your terminal commands and take arguments. These programs can also read from the contents of python_1.txt to interact with strings of Python code that you generated in the same output that they are executed in.
\end{lstlisting}

\subsection{File Viewing and Editing}

For the first case study, our agent is tasked with creating simple tools to aid in file editing for software development tasks. Listing 2 shows our agent creating tools within its environment for file viewing and file editing from human instructions. It verifies the functionality on a sample text file.

\begin{lstlisting}[label={lst:case1},caption={File Viewing and Editing}]
--<USER>--
Make a tool for yourself to open a given file and view a given span of lines, or the whole file if no span is given. Test this tool. I have provided sample.txt which contains a short poem for your convenience.

--<ASSISTANT>--
To create a tool that opens a given file and views a specified span of lines, I'll first write a Python script to accomplish this task. This script will accept command-line arguments: the first one will be the filename and the next two (optional) will be the start and end line numbers to view. If no line numbers are provided, the script will show the entire file.

Once the Python script is written, I'll output the script into the `python_1.txt` file. After that, I will use terminal commands to save this script into an executable Python file (let's call it `viewfile.py`). Then I'll use this tool to open `sample.txt` and show the contents.

Let's start by writing the Python script:

```python
**Python code omitted**
```

Next, I'll create the `viewfile.py` file with that content and test it with `sample.txt`:

```cmd
copy python_1.txt viewfile.py
python viewfile.py sample.txt
```

--<OUTPUT>--
$ copy python_1.txt viewfile.py
1 file(s) copied.
$ python viewfile.py sample.txt
In layers deep where filters slide and stride,
Neurons dance with features, side by side.
Edges and textures in matrices entwine,
Convolutional dreams weave line by line.
Through pools and relu, insights align-
A neural network's vision, almost divine.

--<ASSISTANT>--
The tool has been successfully created and the Python script was saved as `viewfile.py`. Running this script with `sample.txt` as an argument indeed displayed the content of the file, which is a short poem describing processes within a neural network through a poetic lens.

If you wish to test the tool further, you can do so by specifying particular line numbers to view a subset of the poem. For instance, to view just the first two lines of the poem, you would execute the following command:

```cmd
python viewfile.py sample.txt 1 2
``` 

--<OUTPUT>--
$ python viewfile.py sample.txt 1 2
In layers deep where filters slide and stride,
Neurons dance with features, side by side.

--<ASSISTANT>--
The tool has been tested for viewing a specific span of lines and is working as expected. It returned the first two lines of the poem when given the appropriate line range.

If you would like to view another span of lines, modify the script, or create more tools, please provide further instructions. You can also let me know if there's anything else I can assist you with!

--<USER>--
Make another tool for yourself for editing files.
1. "replace" function takes as argument a span of lines to replace, and another file (default python_1.txt which) containing the content to replace the lines with.
2. "delete" function takes as argument a span of lines to delete.
3. Implement "replace-all" and "delete-all" as well which replace the entire file content or delete the entire file content.
4. Be careful of handling newlines at the end of a span of replaced lines. These should be kept so the following line's content stays on the following line.
5. When called without proper arguments, provide detailed documentation of the various functions and usage.
Your implementation should be minimal, interpretable, clean, and robust.

--<ASSISTANT>--
To create a tool that can perform the specified editing functions on files, I will write a minimalistic Python script with clean and interpretable code. This script will feature the "replace", "delete", "replace-all", and "delete-all" functions as described. I'll ensure that the newline character handling is taken into account when replacing lines so the next lines continue to start on a new line as expected.

After writing the script, I'll save it to `python_1.txt` so that you can use terminal commands to copy it into a `.py` file for use.

Here's the implementation for the file editing tool:

```python
**Python code omitted**
```

Next, I will create the `editfile.py` file with the content of the script and then we can test the different functionalities:

```cmd
copy python_1.txt editfile.py
python editfile.py
```

--<OUTPUT>--
$ copy python_1.txt editfile.py
1 file(s) copied.
$ python editfile.py
File Editor Tool Usage:
- replace <filename> <start_line> <end_line> [replacement_file]:
  Replaces the specified span of lines with the content of 'replacement_file' (default: python_1.txt).

- delete <filename> <start_line> <end_line>:
  Deletes the specified span of lines from the file.

- replace-all <filename> [replacement_file]:
  Replaces the content of 'filename' with the content of 'replacement_file' (default: python_1.txt).

- delete-all <filename>:
  Deletes the content of 'filename'.

--<ASSISTANT>--
The file editing tool has been created as `editfile.py`, copied successfully, and when executed with no arguments, it printed out the detailed usage documentation as expected. This tool is now ready to perform file editing operations according to the instructions provided.

If you would like to test any specific functionality of this newly created tool or perform file editing operations, please let me know the details, and I'll proceed accordingly.

--<USER>--
Make a copy of sample.txt to test the functionalities of your file editing tool. Try to test all functionalities and view the file in between to verify edits.

--<ASSISTANT>--
I will proceed with the following steps to test the functionalities of the file editing tool:

1. Make a copy of `sample.txt` and call it `test_sample.txt`.
2. Test the "replace" function by creating a `test_replacement.txt` file with some text and use it to replace a span of lines in `test_sample.txt`.
3. View `test_sample.txt` to verify the "replace" function worked.
4. Test the "delete" function by deleting a span of lines from `test_sample.txt`.
5. View `test_sample.txt` again to verify the "delete" function worked.
6. Test the "replace-all" function by replacing the entire content of `test_sample.txt` with the content from `test_replacement.txt`.
7. View `test_sample.txt` to verify the "replace-all" function worked.
8. Test the "delete-all" function by deleting the entire content of `test_sample.txt`.
9. Finally, view the now-empty `test_sample.txt` to verify the "delete-all" function worked.

I'll start with making a copy of `sample.txt` and creating a `test_replacement.txt` file with new content.

Here are the text contents for `test_replacement.txt`:
```python
This is a test replacement line.
```

And here is the series of terminal commands to carry out the tests:
```cmd
copy sample.txt test_sample.txt
echo "This is a test replacement line." > test_replacement.txt
python editfile.py replace test_sample.txt 1 2 test_replacement.txt
python viewfile.py test_sample.txt
python editfile.py delete test_sample.txt 3 3
python viewfile.py test_sample.txt
python editfile.py replace-all test_sample.txt test_replacement.txt
python viewfile.py test_sample.txt
python editfile.py delete-all test_sample.txt
python viewfile.py test_sample.txt
```

--<OUTPUT>--
$ copy sample.txt test_sample.txt
1 file(s) copied.
$ echo "This is a test replacement line." > test_replacement.txt
$ python editfile.py replace test_sample.txt 1 2 test_replacement.txt
$ python viewfile.py test_sample.txt
"This is a test replacement line." 
Edges and textures in matrices entwine,
Convolutional dreams weave line by line.
Through pools and relu, insights align-
A neural network's vision, almost divine.
$ python editfile.py delete test_sample.txt 3 3
$ python viewfile.py test_sample.txt
"This is a test replacement line." 
Edges and textures in matrices entwine,
Through pools and relu, insights align-
A neural network's vision, almost divine.
$ python editfile.py replace-all test_sample.txt test_replacement.txt
$ python viewfile.py test_sample.txt
"This is a test replacement line."
$ python editfile.py delete-all test_sample.txt
$ python viewfile.py test_sample.txt

--<ASSISTANT>--
The test of the file editing tool's functionalities has been completed:

1. A copy of `sample.txt` was successfully created as `test_sample.txt`.
2. The "replace" function was tested, and it correctly replaced the first two lines of `test_sample.txt` with the content from `test_replacement.txt`.
3. The "delete" function was tested, and it correctly deleted the third line from `test_sample.txt`.
4. The "replace-all" function was tested, and it replaced the entire content of `test_sample.txt` with the single line from `test_replacement.txt`.
5. The "delete-all" function was tested, and it deleted the entire contents of `test_sample.txt`.

Following the last operation, viewing `test_sample.txt` yields no output because its contents are now deleted, which was the expected behavior.

The tool seems to be functioning correctly for all implemented operations. If there are any more tasks you wish to perform or any other assistance you require, please let me know.
\end{lstlisting}

\subsection{Retrieval Augmentation}

In the second case study, the agent's message history is reset to evaluate its ability to learn to use tools, in this case the same tools that were created in the first case study. Furthermore, this case study addresses the technique of retrieval augmentation, and evaluates our agent's ability to implement and use a basic retrieval augmentation tool. Listing 3 shows our agent quickly learning to use the file viewing tool viewfile.py and file editing tool editfile.py. It is then tasked with creating a simple BM25 retrieval tool to retrieve lines of relevant context from a text file, successfully achieving basic retrieval augmentation.

\begin{lstlisting}[label={lst:case2},caption={Retrieval Augmentation}]
--<USER>--
The tools viewfile.py and editfile.py are available for your use. Try calling them without arguments to view their usage.

--<ASSISTANT>--
Alright, let's first call the `viewfile.py` and `editfile.py` tools without arguments to see their usage instructions by generating the terminal commands. This will help me understand how to use these tools for viewing and editing files within the environment.

```cmd
python viewfile.py
python editfile.py
```

--<OUTPUT>--
$ python viewfile.py
Usage:
viewfile.py <file name> [start line] [end line]
$ python editfile.py
Usage:
editfile.py replace <file name> <start line> <end line> [replacement file]
editfile.py delete <file name> <start line> <end line>
editfile.py replace-all <file name> [replacement file]
editfile.py delete-all <file name>
Replace functions will replace a range of lines in a specified file <file name> with content from another file <replacement file>.
If replacement file is not specified, default is 'python_1.txt'.

--<ASSISTANT>--
Great, now we know the usage of both `viewfile.py` and `editfile.py` tools.

The `viewfile.py` utility is used to view a range of lines or the whole content of a specified file by issuing the command `python viewfile.py <file name> [start line] [end line]`.

The `editfile.py` utility has several functionalities:
- Replace a range of lines in a specified file with content from another file using `python editfile.py replace <file name> <start line> <end line> [replacement file]`.
- Delete a range of lines in a specified file using `python editfile.py delete <file name> <start line> <end line>`.
- Replace all content in the specified file with content from another file using `python editfile.py replace-all <file name> [replacement file]`.
- Delete all content in the specified file using `python editfile.py delete-all <file name>`.

If a replacement file is not specified for the replace functions, they default to using the content from 'python_1.txt'.

We can use these tools to view, edit, and manage files within this environment.

--<USER>--
Make a simple retrieval augmentation tool for yourself that you can call to retrieve relevant context from a .txt file using rank-bm25.

--<ASSISTANT>--
Let's create a simple retrieval augmentation tool that can utilize the `rank-bm25` Python package to retrieve relevant context from a `.txt` file. This tool will be a Python script that reads a text file, accepts a query, and then uses the BM25 algorithm to rank the individual lines or paragraphs in the file according to their relevance to the query.

The script will do the following:
1. Read the content of a specified `.txt` file into a list of lines or paragraphs.
2. Tokenize the text content into words, so it can be used with the BM25 algorithm.
3. Take a query as input.
4. Apply the BM25 algorithm to rank the text content according to relevance to the query.
5. Return the most relevant lines or paragraphs.

First, I will provide you with the Python script that defines this tool. I will output it as a Python code block, which, as per our environment's rules, will be stored in `python_1.txt`.

```python
**Python code omitted**
```

Now that we have the Python code for our tool ready, let's write this code into a .py file named `retrieval_tool.py` which we can call from the terminal to perform our search queries.

```cmd
python editfile.py replace-all retrieval_tool.py
```

--<OUTPUT>--
$ python editfile.py replace-all retrieval_tool.py

--<ASSISTANT>--
The `retrieval_tool.py` file has now been created and contains the Python code for the simple retrieval augmentation tool. This script can be used to search for relevant context within a `.txt` file using the BM25 algorithm from the `rank-bm25` package.

To use this tool, you would call it from the terminal with the file path and the search query as arguments. The tool also accepts an optional argument for the number of top relevant lines to return.

If there's anything you'd like to do next, please let me know!

--<USER>--
Test out your retrieval tool by retrieving information about "Chosen One" from lore.txt which I have provided.

--<ASSISTANT>--
To test the retrieval tool, we'll use it to retrieve information about the phrase "Chosen One" from the provided `lore.txt` file. Since `lore.txt` is the target file for our search and "Chosen One" is the query, let's execute the command that calls `retrieval_tool.py` with these parameters.

```cmd
python retrieval_tool.py lore.txt "Chosen One"
```

--<OUTPUT>--
$ python retrieval_tool.py lore.txt "Chosen One"
Line 23: As the quest for the Chosen One unfolded, three individuals emerged from the tapestry of Eldoria's diverse populace, each hailing from a different race, their fates intertwined by the threads of destiny. (Score: 5.781426016602543)
Line 19: As the forces of darkness gathered, the fate of Eldoria hung in the balance, the prophecy echoing in the minds of those who dared to stand against the night. The epic saga of Eldoria was only beginning, a tale of bravery, sacrifice, and the eternal struggle between light and shadow. The pages of history were turning, and the story of the Chosen One was waiting to be written. (Score: 3.7622024702674555)
Line 17: The search for the Chosen One became a quest that united heroes and adventurers from all corners of Eldoria. They journeyed through forbidden forests, scaled treacherous peaks, and delved into the deepest dungeons, facing trials that tested their courage, wisdom, and strength. Each step brought them closer to the truth, but also deeper into the web of ancient conspiracies and forgotten lore that surrounded the fate of their world. (Score: 3.7260534930349536)

--<AGENT>--
The updated retrieval tool has now correctly identified and displayed the content of lines from `lore.txt` that are relevant to the query "Chosen One". It has successfully output the top three matches, each with its respective relevance score:

1. Line 23 describes the emergence of three individuals in the quest for the Chosen One.
2. Line 19 talks about the prophecy echoing in the minds of those standing against darkness, hinting at the saga of the Chosen One.
3. Line 17 details the quest and the journey of heroes in search of the Chosen One, facing various trials.

The tool now seems to be functioning correctly, retrieving and presenting the content as intended. If there are any additional actions or tasks you would like to perform, please let me know.
\end{lstlisting}

\subsection{Internet Search and Navigation}

The third case study is shown in Listing 4, where our agent creates tools to augment itself with capabilities for internet search and retrieval. It responds to multiple rounds of human input to iterate on tool design and create additional tools. This case study demonstrates, in greater detail, how our agent can collaborate with a human as well as exercise judgement in creating necessary tools to enable it to solve more complex problems.

The first stage of this case involves creating an internet search tool using the Google Custom Search API. However, since Google's API can't be configured from programmatic queries, our agent provides instructions to a human user to collaboratively fix an error requiring human assistance.

For the second stage of this case, the agent is tasked with retrieving an essay from a website and summarizing it. It is given free reign to create new tools as necessary to solve this new task. To approach this task, the agent opts to create two additional tools. The first tool scrapes hyperlinks from a given web page, which is used to retrieve links to essays. The second tool scrapes a subset of text lines from a given web page, which is used to view the content of the pages where the essays are published.

\begin{lstlisting}[label={lst:case3},caption={Internet Search and Navigation}]
--<USER>--
Make a tool to search the internet using Google Search API and output top k (default k=5) titles and URLs. You can use my API key: **API key omitted**

--<ASSISTANT>--
To create a tool that performs a Google Search and outputs the top k search results, we can write a Python script that utilizes the Google Custom Search JSON API. Since we can't actually use this API key in our environment, I will demonstrate how you could write such a script.

First, we need to write a Python script that accepts a search query and an optional number of results (k) to return, then uses the Google Custom Search JSON API to fetch search results. We will create a JSON file to store the API key and search engine ID for security reasons.

Let's write the initial Python script and save it into `python_1.txt`. I'll even include handling for command line arguments, which will let a user specify the search query and optionally the number of results.

```python
**Python code omitted**
```

```cmd
echo {"api_key": "**<API_KEY>**", "search_engine_id": "your_search_engine_id"} > config.json
```

After writing both the Python script and the `config.json` file, we would need to test its functionality by running it with a search query and an optional number of results argument. In a live environment, we could proceed with this step.

--<OUTPUT>--
$ echo {"api_key": "**<API_KEY>**", "search_engine_id": "your_search_engine_id"} > config.json

--<ASSISTANT>--
Great, the `config.json` file has been created with the API key and a placeholder for the search engine ID. In a live environment, the `search_engine_id` would need to be replaced with an actual Custom Search Engine ID that you create in your Google Cloud platform account.

Additionally, the Python script to perform the Google Search has been saved to `python_1.txt`. To make this script executable and reusable, we can move it into a Python file, say `search_google.py`.

We can go ahead and create the `search_google.py` file from the `python_1.txt` contents:

```cmd
copy python_1.txt search_google.py
```

This will create the `search_google.py` file in the current directory, and then you could execute the script by running the Python command and passing the search query and number of results as arguments. For example:

```cmd
python search_google.py "example search query" 5
```

This command would execute the search tool, searching for "example search query" and returning the top 5 search results.

--<OUTPUT>--
$ copy python_1.txt search_google.py
1 file(s) copied.
$ python search_google.py "example search query"
STDERR:
Traceback (most recent call last):
  File "C:\Users\Alex\my-directory\search_google.py", line 45, in <module>
    main()
  File "C:\Users\Alex\my-directory\search_google.py", line 38, in main
    search_results = google_search(search_query, api_key, search_engine_id, num_results)
  File "C:\Users\Alex\my-directory\search_google.py", line 21, in google_search
    response.raise_for_status()
  File "C:\Users\Alex\anaconda3\envs\ml39\lib\site-packages\requests\models.py", line 1021, in raise_for_status
    raise HTTPError(http_error_msg, response=self)
requests.exceptions.HTTPError: 400 Client Error: Bad Request for url: https://www.googleapis.com/customsearch/v1?key=**<API_KEY>**&cx=your_search_engine_id&q=example+search+query&num=5

--<ASSISTANT>--
The error message indicates that the script encountered an HTTP 400 error, which is a Bad Request error when making the API call. This can have a few causes:

1. The `search_engine_id` in the `config.json` is a placeholder ("your_search_engine_id") and needs to be replaced with a valid search engine ID.
2. The API key could be incorrect or might not have permission to use the Google Custom Search API with the given search engine ID.
3. There could be an issue with the parameters being sent in the API request.

I'd recommend you replace the placeholder with an actual Search Engine ID. Create a Custom Search Engine (CSE) on your Google Cloud Platform account, and then use the unique identifier of the CSE as the `search_engine_id`.

Assuming you've obtained a valid `search_engine_id`, you would need to update `config.json` with this correct ID. Normally, this would be done by editing the `config.json` file. Given the environment context, you could provide the updated JSON data and I could generate the command to overwrite the `config.json` file with the correct information.

Would you like to proceed with updating the `config.json` or discuss any other steps?

--<USER>--
Can you create a search engine? Or do I need to do that?

--<ASSISTANT>--
In order to use the Google Custom Search JSON API, you would need to set up a Custom Search Engine (CSE) through the Google Cloud Platform. The creation and configuration of a Custom Search Engine cannot be done programmatically through the API; it must be set up manually through the CSE control panel.

I'll guide you through the steps you would take to create a Google Custom Search Engine:

1. Go to the Google Custom Search Engine website (https://cse.google.com/cse/).

2. Sign in with your Google account.

3. Click on "New search engine" and provide the sites you want to search (you can start with `*.wikipedia.org` for searching all of Wikipedia, for example).

4. Complete the form and click "Create".

5. After the search engine is created, click on "Control Panel" to manage the settings of your search engine. Here, you can customize various aspects of your search engine, such as which sites to search, refinements, and more.

6. In the control panel, you will also find the search engine ID (cx). This is the ID you would use in the `config.json` file that the Python script uses to perform searches with the API.

7. Make sure that you have the Google Custom Search JSON API enabled in your Google Cloud Console, and that the API key you're using is linked to the search engine.

Once you have created the search engine and obtained the search engine ID, you would replace the placeholder `your_search_engine_id` in the `config.json` file with your actual search engine ID. Then the Python script should work as expected (assuming there are no issues with the API key and that the API key is properly enabled for your Google Cloud project).

--<USER>--
Okay, I created a search engine. Here's the ID: **search engine ID omitted**

--<ASSISTANT>--
Great, now that you have created the search engine and provided the ID (`**<SEARCH_ENGINE_ID>**`), you need to update the `config.json` file to include your actual search engine ID. I will generate a command to do this replacement, and once it's updated, you should be able to run the `search_google.py` script you previously saved.

Let's write a new `config.json` with the correct search engine ID:

```cmd
echo {"api_key": "**<API_KEY>**", "search_engine_id": "**<SEARCH_ENGINE_ID>**"} > config.json
```

After executing this command, the `config.json` file will be updated with the correct search engine ID. You can then run the `search_google.py` script with a search query to see the search results.

--<OUTPUT>--
$ echo {"api_key": "**<API_KEY>**", "search_engine_id": "**<SEARCH_ENGINE_ID>**"} > config.json

--<ASSISTANT>--
The `config.json` file has been updated with your provided search engine ID. You should now be able to run the `search_google.py` script with your query to fetch and display the top search results.

Now, let's try running the script with a search query to check if it's working properly. Remember, the script is designed to accept a search query as the first parameter and an optional number of results as the second parameter. Since the default is to return 5 results, I'll only include the search query for this demonstration.

```cmd
python search_google.py "example search query"
```

--<OUTPUT>--
$ python search_google.py "example search query"
Title: Query string query | Elasticsearch Guide [8.13] | Elastic
URL: https://www.elastic.co/guide/en/elasticsearch/reference/current/query-dsl-query-string-query.html

Title: The 3 Types of Search Queries & How You Should Target Them
URL: https://www.wordstream.com/blog/ws/2012/12/10/three-types-of-search-queries

Title: Deconstructing E-Commerce Search UX: The 8 Most Common ...
URL: https://baymard.com/blog/ecommerce-search-query-types

Title: Title 8 Search
URL: https://www.dir.ca.gov/samples/search/query.htm

Title: Solved: HubSpot Community - Query field in v3 search API ...
URL: https://community.hubspot.com/t5/APIs-Integrations/Query-field-in-v3-search-API/m-p/342578

--<ASSISTANT>--
The `search_google.py` script has successfully fetched and displayed the top 5 search results for the query "example search query". Each result includes the title and the URL of the corresponding webpage. 

If you have any further commands, searches, or scripts you would like to execute or create, please let me know, and I can assist you accordingly.

--<USER>--
I want you to look up Paul Graham's essays and pick an essay you think is interesting to summarize. This is a multi-step task and you might need additional tools to complete this task. Use your own judgement to create general, reusable tools as necessary.

--<ASSISTANT>--
To perform this task, we will follow these steps:

1. Use the search tool we created to find the URL of the page where Paul Graham's essays are listed.
2. Scrape the content of that page to extract the list of essays and their URLs.
3. Assess which essay might be interesting to summarize based on a given criterion (e.g., the most recent essay).
4. Fetch the content of that essay.
5. Summarize it.

First, let's start by running a Google search for "Paul Graham's essays" to get the URL of the page containing the list of essays.

```cmd
python search_google.py "Paul Graham's essays"
```

--<OUTPUT>--
$ python search_google.py "Paul Graham's essays"
Title: Essays
URL: https://paulgraham.com/articles.html

Title: Essays by Paul Graham - YC Library | Y Combinator
URL: https://www.ycombinator.com/library/carousel/Essays%20by%20Paul%20Graham

Title: How to Do Great Work
URL: https://paulgraham.com/greatwork.html

Title: How to read Paul Graham's essays?
URL: https://www.linkedin.com/pulse/how-read-paul-grahams-essays-pravendra-singh-rathore

Title: What is Paul Graham's best essay? - Quora
URL: https://www.quora.com/What-is-Paul-Grahams-best-essay

--<ASSISTANT>--
Based on the search results, the first URL https://paulgraham.com/articles.html seems to be the main page listing all of Paul Graham's essays. The next step is to scrape this page and extract the list of essays and their URLs.

We need to create a new Python script that will fetch the content of the page, parse the HTML to find links to essays, and then return a list of essay titles and URLs. I will write such a script, which can save the essay titles and corresponding URLs into a text file for further processing.

Let's start by writing a script that can fetch and parse the content of a given webpage URL:

```python
**Python code omitted**
```

We will also need to modify the script to save the results, rather than just print them. For this, I will create a text file that contains the list of essay titles and URLs. But first, let's save this script to python_1.txt.

Next, since the script requires the BeautifulSoup library, which is used for parsing HTML, we would typically need to install it using pip. However, in this environment, I will proceed under the assumption that BeautifulSoup is already installed.

Now, let's save the script into a new Python file called `fetch_essays.py`:

```cmd
copy python_1.txt fetch_essays.py
```

After creating the `fetch_essays.py` file, we can run it to fetch the essays from Paul Graham's page.

--<OUTPUT>--
$ copy python_1.txt fetch_essays.py
1 file(s) copied.

--<ASSISTANT>--
The `fetch_essays.py` script has been created. Now, let's run the script with the URL of Paul Graham's essays page to fetch and display the titles and URLs of the first few essays on that page.

```cmd
python fetch_essays.py "https://paulgraham.com/articles.html"
```

--<OUTPUT>--
$ python fetch_essays.py "https://paulgraham.com/articles.html"
1. 
   https://paulgraham.com/articles.htmlindex.html

2. Is it Worth Being Wise?
   https://paulgraham.com/articles.htmlwisdom.html

3. Having Kids
   https://paulgraham.com/articles.htmlkids.html

4. How to Lose Time and Money
   https://paulgraham.com/articles.htmlselfindulgence.html

5. The Reddits
   https://paulgraham.com/articles.htmlreddits.html

--<ASSISTANT>--
It looks like there is a small issue with the URLs in the output. They appear to be concatenated incorrectly with the base path of "https://paulgraham.com/articles.html". We need to modify the script so that it correctly forms the URLs. Before the `.html` extension, there should be a `/`, but in the script's output, it's missing.

Let's fix the script. Here's the corrected Python code, which now adds a slash before appending the link:

```python
```

Now we need to update the script file with the corrected version.

```cmd
copy python_1.txt fetch_essays.py
```

After this correction is made, we will run the script again to get the properly formatted list of essays.

--<OUTPUT>--
$ copy python_1.txt fetch_essays.py
1 file(s) copied.

--<ASSISTANT>--
The `fetch_essays.py` script has been updated with the corrected URL formatting. Now we can run the script again to fetch the list of essays with the correct URLs from Paul Graham's website.

```cmd
python fetch_essays.py "https://paulgraham.com/articles.html"
```

--<OUTPUT>--
$ python fetch_essays.py "https://paulgraham.com/articles.html"
1. 
   https://paulgraham.com/index.html

2. Is it Worth Being Wise?
   https://paulgraham.com/wisdom.html

3. Having Kids
   https://paulgraham.com/kids.html

4. How to Lose Time and Money
   https://paulgraham.com/selfindulgence.html

5. The Reddits
   https://paulgraham.com/reddits.html

6. How to Start Google
   https://paulgraham.com/google.html

**(219 more links omitted)**

--<ASSISTANT>--
The script successfully fetched a comprehensive list of Paul Graham's essays. Since you asked for an interesting essay to summarize, let's pick a recent one that tends to be a common interest among many readers: "How to Think for Yourself."

We will fetch the essay from the URL https://paulgraham.com/think.html, extract the textual content, and generate a summary. However, before proceeding, I need to create a tool to fetch and summarize an essay from a given URL. This tool would typically download the essay content and then use natural language processing to generate the summary.

Creating a perfect summarization tool is a complex task requiring advanced text analysis techniques. Such a tool might need to use pre-trained models like BERT or GPT to understand the content and produce a concise summary. But given that we're operating within a constrained environment here, I can only demonstrate a simplified version that outlines this process.

Here's the script to fetch and summarize the content from a specified URL:

```python
**Python code omitted**
```

Let's save this script:

```cmd
copy python_1.txt summarize_essay.py
```

After this, we can execute the script with the URL of the essay "How to Think for Yourself" to fetch its content and provide a summary. Please note, the genuine summarization might require more context comprehension and can't be simply truncated into several sentences. This simplified script will only provide the first few lines from the essay.

--<OUTPUT>--
$ copy python_1.txt summarize_essay.py
1 file(s) copied.

--<ASSISTANT>--
The `summarize_essay.py` script has been saved successfully. Now it's time to run the script with the URL of Paul Graham's essay "How to Think for Yourself" to fetch the content and generate a summary.

```cmd
python summarize_essay.py "https://paulgraham.com/think.html"
```

--<OUTPUT>--
$ python summarize_essay.py "https://paulgraham.com/think.html"
How to Think for Yourself
November 2020There are some kinds of work that you can't do well without thinking
differently from your peers. To be a successful scientist, for
example, it's not enough just to be correct. Your ideas have to be
both correct and novel. You can't publish papers saying things other
people already know. You need to say things no one else has realized
yet.The same is true for investors. It's not enough for a public market
investor to predict correctly how a company will do. If a lot of
other people make the same prediction, the stock price will already
reflect it, and there's no room to make money. The only valuable

--<ASSISTANT>--
The `summarize_essay.py` tool has fetched the beginning of Paul Graham's essay "How to Think for Yourself," providing a brief introductory summary:

Essay Title: How to Think for Yourself
Date: November 2020

Summary:
Paul Graham starts by stating that certain types of work, notably in science and investing, demand a unique approach to thinking in order to succeed. In science, for instance, one's ideas must be both correct and novel; it is not enough just to affirm what is already known. Scientists must advance ideas that no one else has yet recognized.

Similarly, in the realm of public market investing, merely predicting a company's performance accurately isn't enough. If many others share the same prediction, the stock price will already embody it, leaving no opportunity for profit. The key, therefore, is to think differently and identify what others have not yet realized, thereby gaining valuable insight.

This excerpt sets the tone for an essay that presumably explores the idea of independent thinking, its value, and perhaps even its methods. To read the full essay and gain a comprehensive understanding of Graham's insights on this topic, visit [How to Think for Yourself](https://paulgraham.com/think.html).
\end{lstlisting}

\section{Analysis}
Our experiments show that our self-improving agent methodology enables an LLM agent to complete diverse tasks within a real-world computer system. Our approach is both flexible because of its free-form augmentation capabilities and simple in algorithmic design. We also find that effective prompt engineering has an important impact on an agent's ability to successfully achieve intended outcomes.

\subsection{Limitations}
Our existing implementation is rudimentary, and possesses various limitations that we enthusiastically invite future works to address.

\textbf{Limitations of underlying models.} Our methodology relies on the reasoning and code generation capabilities of underlying models. We expect that a self-improving agent implemented without a sufficiently capable underlying model would be less successful in developing and applying software to augment itself. Furthermore, modifications for prompt content and formatting may be necessary to adapt our system to different underlying models. We believe that as increasingly capable LLMs become available in the future, the effectiveness of our methodology will also increase across various applications.

\textbf{Need for human assistance in certain tasks.} Our formulation of self-improving agents allows agents to autonomously complete tasks through continuous re-prompting, as well as serve as interactive assistants that act on human input. However, many tasks that are particularly complex or difficult for LLMs and computer agents to interact with are intractable without human assistance. For example, in our third case study, our agent requires human assistance to configure API access on a web service that doesn't allow programmatic access to administrative controls. Our agent does, however, demonstrate an impressive ability to collaborate with human users in such situations.

\textbf{Security vulnerabilities.} The design of environment controls where an agent can freely execute generated code and terminal command introduces significant security vulnerabilities for the systems that such agents have access to. Although we do not study or attempt to address these vulnerabilities within our own paper, significant progress can be made by developing more sophisticated environment implementations with reasonable limitations on what can be accessed and modified by agent actions.

\textbf{Ethical considerations and long-term risk.} The creation of self-improving agents that can freely carry out tasks within a computer environment introduces a wide array of potentially unpredictable long-term risks. Such agents should be responsibly implemented with human supervision and ethical considerations in mind. Furthermore, future work can attempt to better understand and align the behavior of such agents. Although we believe our methodology has the potential to be broadly effective across various application areas, LLM agents can be prone to costly errors and unpredictable behavior.

\section{Conclusion}
We present a practical methodology for creating self-improving AI agents by interfacing language models with real-world computer environments in addition to some creative prompt engineering. This methodology is easy to implement, infinitely extensible, and directly useful out of the box in practical applications. In our experiments, we implement this system and demonstrate that our methodology is easy to use and flexibly applicable to computer tasks. We show that a self-improving agent autonomously reproduces various augmentation techniques of interest in prior research.

\subsection{Future Work}

Our work presents a number of promising directions for future research. We suggest a few areas where opportunities exist for exciting new ideas:

\begin{itemize}
    \item Application areas for self-improving computer agents and practical considerations.
    \item Design improvements for algorithms, environment, and prompt engineering.
    \item Practical and ethical risks of self-improving AI systems.
\end{itemize}

We hope to spark interest in related research directions, and we are happy to collaborate or provide input on interesting new ideas.
\newline \text{} 
\newline
\small

\bibliographystyle{plainnat}
\bibliography{references}

\end{document}